%% file: main.tex
\documentclass{article}

\usepackage{arxiv}
\usepackage{amsmath}
\usepackage{amssymb}
\usepackage{graphicx}
\usepackage{subcaption}
\usepackage{hyperref}
\usepackage{enumitem}
\usepackage{float}
\usepackage{placeins}
\usepackage{booktabs}
\usepackage{array}
\setlist{nosep}
\title{Sentiment Analysis of Mobile Legends App Reviews Using Machine Learning and LSTM-Based Deep Learning Models}

\author{
Vira Putri Maharani \\
Department of Data Science \\
Institut Teknologi Sumatera \\
Lampung Selatan, Indonesia \\
\texttt{vira.122450129@student.itera.ac.id}
\And
Kharisa Harvanny \\
Department of Data Science \\
Institut Teknologi Sumatera \\
Lampung Selatan, Indonesia \\
\texttt{kharisa.122450061@student.itera.ac.id}
\AND
Daris Samudra \\
Department of Data Science \\
Institut Teknologi Sumatera \\
Lampung Selatan, Indonesia \\
\texttt{daris.122450102@student.itera.ac.id}
\And
Luluk Muthoharoh, M.S \\
Department of Data Science \\
Institut Teknologi Sumatera \\
Lampung Selatan, Indonesia \\
\texttt{luluk.muthoharoh@sd.itera.ac.id}
\AND
Ardika Satria, M.Si \\
Department of Data Science \\
Institut Teknologi Sumatera \\
Lampung Selatan, Indonesia \\
\texttt{ardika.satria@sd.itera.ac.id}
\And
Martin Clinton Tosima Manullang, Ph.D. \\
Department of Informatics Engineering \\
Institut Teknologi Sumatera \\
Lampung Selatan, Indonesia \\
\texttt{martin.manullang@if.itera.ac.id}
}

\begin{document}
\fontsize{8}{9.2}\selectfont
\maketitle

\begin{abstract}
Mobile game reviews often contain informal language, slang, abbreviations, and mixed-language expressions, making sentiment classification a challenging task. This study compares Machine Learning (ML) models and Deep Learning (DL) using Long Short-Term Memory (LSTM) for sentiment analysis on Mobile Legends app reviews. The dataset consists of 10,000 user reviews categorized into positive, negative, and neutral sentiments. The ML approach uses TF-IDF and PyCaret AutoML to evaluate multiple models, while the DL approach uses an LSTM architecture to capture sequential dependencies. The results show that LSTM achieves an accuracy of 92 and a weighted F1-score of 91, outperforming traditional ML models. These findings highlight the effectiveness of deep learning in handling sequential text data. Source code is publicly available  at  https://github.com/Viramhrani/pba2026-Kelompok16
\end{abstract}
\input{introduction}
\input{related_work}

\input{dataset}
\input{methodology}
\input{result}
\input{deep}

\input{archi}
\input{train}
\input{know}
\input{references}

\end{document}

%% file: introduction.tex
\section{Introduction}
The global mobile gaming industry has experienced rapid growth over the past decade, with Southeast Asia emerging as one of the fastest-growing markets. Among the most popular titles in this region is Mobile Legends: Bang Bang (MLBB), developed by Moonton. As a multiplayer online battle arena (MOBA) game, MLBB attracts millions of active players and generates a continuous stream of user reviews on platforms such as the Google Play Store. These reviews cover diverse aspects, including gameplay experience, graphical improvements, server performance, cheating issues, and feature requests.\\
This large volume of user-generated content represents a valuable yet underutilised source of user feedback. Manually analysing thousands of reviews on a daily basis is impractical for development teams. Therefore, sentiment analysis—defined as the automated process of identifying and classifying opinions expressed in text—provides a scalable and efficient solution. Recent studies highlight that sentiment analysis enables organisations to systematically extract insights from large-scale textual data and monitor user satisfaction trends over time Sentiment Analysis (Jim et al., 2024; Alharbi et al., 2024). By categorising reviews into positive, neutral, and negative sentiments, developers can prioritise bug fixes, evaluate feature updates, and improve overall user experience.\\

However, performing sentiment analysis on Indonesian-language reviews presents unique challenges compared to English-based natural language processing tasks. The Indonesian language exhibits rich morphological structures, where a single root word can generate multiple surface forms through affixation, including prefixes (e.g., me-, ber-, di-) and suffixes (e.g., -kan, -an, -i). In addition, user-generated reviews often contain informal expressions, slang, abbreviations, and code-switching between Indonesian and English. These characteristics significantly increase vocabulary variability and can negatively impact classification performance if not handled properly. Recent research emphasises the importance of language-specific preprocessing techniques, such as stemming and stopword removal, to address these challenges and improve model accuracy (Islam et al., 2024).\\
Previous studies have explored various approaches for sentiment analysis in Indonesian text, ranging from classical machine learning algorithms such as Logistic Regression, Naive Bayes, and Random Forest to deep learning models including Long Short-Term Memory (LSTM) and Transformer-based architectures like IndoBERT. While these approaches have shown promising results, many studies primarily focus on maximising accuracy without considering practical deployment constraints. In real-world applications, models must be computationally efficient, lightweight, and reproducible, especially when deployed on limited-resource environments such as free-tier cloud platforms.
Moreover, there is still limited research that directly compares automated machine learning frameworks, such as PyCaret, with custom deep learning implementations under consistent preprocessing settings on the same dataset. This gap highlights the need for a systematic and fair comparison between classical and deep learning approaches in practical sentiment analysis scenarios.\\
This study addresses the gap by conducting an end-to-end comparison between classical machine learning (ML) and LSTM-based deep learning (DL) methods for three-class sentiment classification of Indonesian MLBB reviews. The proposed preprocessing pipeline incorporates Indonesian-specific text normalisation, including stemming and stopword filtering. Classical models are evaluated using PyCaret AutoML with 5-fold cross-validation, while the deep learning model is implemented as a compact LSTM in PyTorch with fewer than 10 million parameters. Both approaches are further deployed as interactive applications using Gradio on Hugging Face Spaces to ensure accessibility and reproducibility.\\
This study aims to answer the following research questions:
\begin{itemize}[leftmargin=*]
\item \textbf How do classical ML classifiers perform on Indonesian MLBB sentiment classification using an AutoML framework?
\item \textbf Can a compact LSTM model outperform classical baselines under identical preprocessing conditions?
\item \textbf Which approach is more suitable for lightweight and interactive deployment?\\
\end{itemize}

The main contributions of this paper are:

\begin{itemize}[leftmargin=*]
\item \textbf{Preprocessing pipeline:} A reproducible Indonesian text cleaning pipeline combining case folding, regex-based noise removal, NLTK stopword filtering, and Sastrawi morphological stemming, with token truncation at 100 tokens.

\item \textbf{ML benchmark:} A PyCaret AutoML comparison of three classifiers (LR, NB, RF) with TF-IDF features and 5-fold stratified cross-validation.

\item \textbf{DL model:} A custom PyTorch LSTM with an embedding layer, single LSTM layer (64 hidden units), dropout regularisation, and a fully-connected head---totalling approximately 690K parameters.

\item \textbf{Deployment:} Two publicly accessible interactive demos on Hugging Face Spaces built with Gradio.

\item \textbf{Open source:} All code, data, and notebooks are publicly available at \url{https://github.com/Viramhrani/pba2026-Kelompok16}.
\end{itemize}

%% file: related_work.tex
\section{Related Work}
\subsection{Machine Learning Models}
Machine Learning (ML) models have been widely adopted in sentiment analysis tasks due to their simplicity and computational efficiency. Classical approaches such as Naive Bayes, Logistic Regression, and Support Vector Machine (SVM) rely on feature extraction techniques such as Bag of Words (BoW) and Term Frequency-Inverse Document Frequency (TF-IDF) to represent text as numerical vectors. These methods treat each word independently, which limits their ability to capture sequential or contextual relationships between words.
Studies on Indonesian-language text classification have demonstrated that Logistic Regression and SVM produce competitive results when combined with appropriate preprocessing and feature engineering [Setyani et al., 2026;Pangestu et al., 2025]. PyCaret, an AutoML framework, has been used to streamline model comparison and evaluation across multiple classifiers in a reproducible manner. Among evaluated models, Random Forest has been shown to achieve the best results in several text classification benchmarks [Ningsih et al., 2024].

\subsection{Deep Learning Approach}
Deep learning models, particularly Long Short-Term Memory (LSTM) networks, have demonstrated superior performance in sentiment analysis of sequential text data. LSTM is a variant of Recurrent Neural Network (RNN) designed to overcome the vanishing gradient problem and to capture long-term dependencies between words in a sentence [11]. Its memory cell, input gate, forget gate, and output gate allow the model to selectively retain or discard information across time steps.
Several studies have confirmed that LSTM outperforms traditional ML models in text classification tasks involving informal and context-dependent language. For instance, research on Indonesian social media comments has shown that LSTM more effectively handles slang, abbreviations, and code-switching compared to Naive Bayes or Logistic Regression [9]. Furthermore, LSTM-based models have been applied to review classification on platforms such as Google Play Store and YouTube, consistently achieving high accuracy and F1-scores [7, 8].

\subsection{Comparison Between ML and DL Approaches}
Machine Learning and Deep Learning approaches have been widely applied in sentiment analysis tasks, each offering different advantages depending on the characteristics of the data. Traditional Machine Learning models such as Naive Bayes, Logistic Regression, and Random Forest rely on feature extraction techniques like TF-IDF and treat text data as independent observations. These models are efficient and relatively simple to implement, but they have limitations in capturing contextual and sequential relationships between words. In contrast, Deep Learning models such as Long Short-Term Memory (LSTM) are specifically designed to process sequential data. LSTM is capable of capturing both short-term and long-term dependencies within text, allowing it to better understand contextual meaning in user-generated content. This makes LSTM particularly effective for handling informal text data, such as mobile game reviews that often contain slang, abbreviations, and inconsistent structures. Comparative studies have shown that deep learning models generally outperform traditional machine learning approaches in text classification tasks, especially when dealing with sequential and context-dependent data. However, machine learning models remain competitive in terms of computational efficiency and are still effective when combined with appropriate feature engineering techniques. Based on these considerations, this study compares Machine Learning models and an LSTM-based Deep Learning model to evaluate their performance in sentiment classification on Mobile Legends user reviews.

%% file: DATASET.tex
\section{Dataset}
\subsection{Mobile Legend App Reviews}

Mobile Legend App Reviews \cite{ref4} is distributed as a semicolon-delimited CSV file encoded in UTF-8, available at 
\url{https://www.kaggle.com/code/auliyyaaini/sentiment-analysis-mobile-legends-app?select=mobile_legends_reviews_cleaned1.csv}. 

The file is stored in this project as 
\texttt{data/MobileLegend\_Apps\_Review\_Dataset.csv}. 

The primary text column is Mobile Legends user review. The class distribution across all 10,000 samples.
\subsection{Data Distribution}
The class distribution of the dataset is highly imbalanced, with negative reviews dominating the dataset. The distribution is as follows: Negative: 6,747 samples (67.47), Positive: 2,373 samples (23.73), and Neutral: 880 samples (8.80). This imbalanced distribution poses a challenge for classification models, particularly for minority classes such as Neutral. The exploratory data analysis further reveals that the dataset contains noise such as slang, emojis, repeated characters, and inconsistent formatting, which necessitates thorough preprocessing before modelling.
\begin{table}[h!]
\centering
\caption{Dataset Summary}
\begin{tabular}{lcc}
\toprule
\textbf{Sentiment Class} & \textbf{Number of Samples} & \textbf{Percentage (\%)} \\
\midrule
Negative & 6,747 & 67.47 \\
Positive & 2,373 & 23.73 \\
Neutral  & 880   & 8.80  \\
\textbf{Total} & \textbf{10,000} & \textbf{100.00} \\
\bottomrule
\end{tabular}
\label{tab:dataset_summary}
\end{table}

%% file: methodology.tex
\section{Methodology}
This study employs a quantitative approach using Natural Language Processing (NLP) methods to classify sentiments from Mobile Legends app reviews. The research pipeline consists of data collection, preprocessing, feature representation, model training, and evaluation.

\subsection{Data Preprocessing}
The preprocessing stage aims to clean and standardize raw text data before modelling. The following steps are performed sequentially:

\begin{enumerate}
    \item \textbf{Case Folding}: All text is converted to lowercase to ensure uniformity.
    \item \textbf{Text Cleaning}: Punctuation, emojis, URLs, mentions, hashtags, numbers, and special characters are removed using regular expressions.
    \item \textbf{Normalization}: Non-standard words and slang are mapped to their standard Indonesian equivalents.
    \item \textbf{Tokenization}: Sentences are split into individual tokens (words).
    \item \textbf{Stopword Removal}: Common Indonesian and English stopwords are removed using the NLTK stopword list.
    \item \textbf{Stemming}: Affixed words are reduced to their root forms using the Sastrawi library for Indonesian morphological processing.
    \item \textbf{Token Truncation}: All sequences are truncated to a maximum of 100 tokens for consistency.
\end{enumerate}

\subsection{Feature Representation}
Two feature representation methods are used depending on the model type.

For Machine Learning models, the preprocessed text is transformed into numerical vectors using Term Frequency-Inverse Document Frequency (TF-IDF). The TF-IDF weight of a term $t$ in document $d$ is defined as:

\[
\text{TF-IDF}(t, d) = tf(t, d) \times \log\left(\frac{N}{df(t)} + 1\right)
\]

where $N$ is the total number of documents and $df(t)$ is the document frequency of term $t$.

For the Deep Learning model, each word is mapped to an integer index and subsequently encoded into dense vector representations via an embedding layer.

\subsection{Classification Models}
This study evaluates four classification models:

\begin{itemize}
    \item \textbf{Naive Bayes (NB)}: A probabilistic classifier based on Bayes' Theorem that assumes feature independence.
    \item \textbf{Logistic Regression (LR)}: A linear model that estimates class probabilities using a logistic function.
    \item \textbf{Random Forest (RF)}: An ensemble method that combines multiple decision trees to reduce overfitting and improve performance.
    \item \textbf{Long Short-Term Memory (LSTM)}: A deep learning architecture that captures sequential and contextual relationships in text data.
\end{itemize}

Machine Learning models (NB, LR, RF) are implemented using the PyCaret AutoML framework with TF-IDF features. The LSTM model is implemented using PyTorch.

\subsection{LSTM Model Architecture}
The LSTM model architecture consists of the following layers:

\begin{itemize}
    \item \textbf{Embedding Layer}: Converts each input token into a dense 64-dimensional vector representation. This allows the model to learn semantic relationships between words.
    \item \textbf{LSTM Layer}: A single LSTM layer with 64 hidden units that processes the embedded word sequences and captures short- and long-term dependencies.
    \item \textbf{Dropout Layer}: Applied after the LSTM layer with a dropout rate of 0.3 to prevent overfitting during training.
    \item \textbf{Fully Connected Layer}: Maps the extracted features to the number of output classes (3: positive, negative, neutral).
    \item \textbf{Output Layer}: Uses the Softmax activation function to produce class probability distributions.
\end{itemize}

\begin{table}[h!]
\centering
\caption{LSTM Hyperparameters}
\begin{tabular}{|l|c|}
\hline
\textbf{Hyperparameter} & \textbf{Value} \\ \hline
Embedding Dimension & 64 \\ \hline
Hidden Dimension & 64 \\ \hline
Total Parameters & $\sim$690K \\ \hline
Batch Size & 32 \\ \hline
Learning Rate & 0.001 \\ \hline
Epochs & 20 \\ \hline
Dropout Rate & 0.3 \\ \hline
Output Classes & 3 \\ \hline
Optimizer & Adam \\ \hline
Loss Function & CrossEntropyLoss \\ \hline
\end{tabular}
\label{tab:lstm_hyperparameters}
\end{table}

\subsection{Evaluation Metrics}
Model performance is evaluated using four standard classification metrics:

\textbf{Accuracy}: The proportion of correctly classified samples out of all predictions.
\[
\text{Accuracy} = \frac{TP + TN}{TP + TN + FP + FN}
\]

\textbf{Precision}: The proportion of positive predictions that are actually correct.
\[
\text{Precision} = \frac{TP}{TP + FP}
\]

\textbf{Recall}: The model's ability to identify all actual positive samples.
\[
\text{Recall} = \frac{TP}{TP + FN}
\]

\textbf{F1-Score}: The harmonic mean of precision and recall.
\[
\text{F1-Score} = 2 \times \frac{\text{Precision} \times \text{Recall}}{\text{Precision} + \text{Recall}}
\]

All models are evaluated on a held-out test set derived from an 80:20 train-test split to ensure that reported metrics reflect generalization performance.

%% file: result.tex
\section{Macine Learning Models}
\subsection{Naive Bayes}
Naive Bayes is a probabilistic classification method based on Bayes’ Theorem. It assumes independence between features, making it simple and efficient for text classification. However, this assumption limits its ability to capture complex relationships in the data. 
\subsection{Logistic Regression}
Logistic Regression is a linear classification model that estimates the probability of a class using a logistic function. Compared to Naive Bayes, this model is more capable of capturing relationships between features, but it still has limitations in handling complex patterns. 
\subsection{Random Forest}
Random Forest is an ensemble learning method that combines multiple decision trees to improve prediction performance and reduce overfitting. In this study, Random Forest achieves the best performance among all machine learning models. 
All machine learning models are implemented using the PyCaret framework, which enables efficient model comparison and evaluation. 

%% file: deep.tex
\section{Deep Learning Models}
\subsection{Long Short-Term Memory (LSTM)}
The deep learning approach in this study uses a Long Short-Term Memory (LSTM) network. The model architecture consists of an embedding layer, an LSTM layer with 50 units, and a dense output layer for classification.  The model is trained using the Adam optimizer and cross-entropy loss function. LSTM is selected because it is capable of capturing sequential dependencies and contextual relationships in text data. This makes it more effective than traditional machine learning models for handling user-generated text such as mobile game reviews.

%% file: archi.tex
\section{Architectural Frameworks}
\subsection{Machine Learning Approach}
The Machine Learning approach in this study follows a structured pipeline consisting of text vectorization, model training, and evaluation using an AutoML framework.\\

First, the preprocessed text data is transformed into numerical features using the Term Frequency–Inverse Document Frequency (TF-IDF) representation. TF-IDF converts textual data into a weighted vector space, where each term reflects its importance within a document relative to the entire corpus. \\

The TF-IDF formulation is defined as:

\begin{equation}
TF\text{-}IDF(t,d) = tf(t,d) \log \left( \frac{N}{df(t)+1} \right)
\end{equation}

where $t$ denotes a term, $d$ represents a document, $N$ is the total number of documents, and $df(t)$ is the document frequency of term $t$.

After vectorization, the dataset is processed using the PyCaret AutoML framework to automate model training, comparison, and evaluation. PyCaret standardizes the experimental pipeline, including data preprocessing, model selection, and cross-validation.\\

In this study, three classification algorithms are evaluated:
\begin{itemize}[leftmargin=*]
\item \textbf{Naive Bayes (NB)}
\item \textbf{Logistic Regression (LR)}
\item \textbf{Random Forest (RF)}
\end{itemize}

All models are trained using TF-IDF features and evaluated using stratified k-fold cross-validation to ensure reliable performance comparison across imbalanced classes.\\

The best-performing model is then selected based on evaluation metrics such as accuracy, precision, recall, and F1-score. The selected model, along with the TF-IDF vectorizer, is saved for deployment purposes. \\

This pipeline ensures that the Machine Learning approach is reproducible, efficient, and suitable for real-world deployment scenarios. Furthermore, the use of PyCaret simplifies experimentation while maintaining consistency across model evaluation.
\subsection{Deep Learning Approaches: LSTM}

The deep learning approach in this study utilizes a Long Short-Term Memory (LSTM) network to perform sentiment classification on Mobile Legends user reviews. LSTM is a variant of Recurrent Neural Networks (RNN) designed to address the vanishing gradient problem and effectively capture long-term dependencies in sequential data. 

The implemented model consists of three main components. First, an embedding layer is used to transform each token into a dense vector representation, allowing the model to learn semantic relationships between words.

Let the embedding matrix be defined as 
\[
E \in \mathbb{R}^{|V| \times d},
\]
where $|V|$ represents the vocabulary size and $d$ denotes the embedding dimension. Each input sequence is converted into a sequence of embedding vectors. All input sequences are padded or truncated to a fixed length $L$ to ensure consistent input dimensions across the dataset.

The sequence of embeddings is then processed by an LSTM layer with 50 hidden units, which captures contextual information across time steps.

The LSTM computation at each time step $t$ is defined as:
\begin{equation}
h_t = \mathrm{LSTM}(x_t, h_{t-1})
\end{equation}
where $x_t$ is the input embedding at time step $t$, and $h_t$ is the hidden state representing contextual information up to time step $t$.

The final hidden representation is passed to a fully connected dense layer, followed by a softmax activation function to produce class probabilities for sentiment classification. The model is trained using the Adam optimizer with categorical cross-entropy as the loss function. During training, the model shows a steady decrease in loss, indicating effective learning of patterns in the data.

Overall, the LSTM model is capable of capturing both short-term and long-term dependencies in textual data, making it well-suited for sentiment analysis of user-generated content such as mobile game reviews.

%% file: train.tex
\section{Training and Evaluation Framework}

\subsection{Training Configuration}

All training experiments are executed through the \texttt{src/train.py} script using a standardized set of hyperparameters across all architectures. This uniform configuration is intended to preserve experimental consistency, thereby enabling a more rigorous and methodologically sound comparison of model performance.

\textbf{Data Split:} The dataset is partitioned into training, validation, and test subsets with an 80:20 ratio. This split allows the model to learn patterns from the majority of the data while reserving a portion for performance evaluation.

\textbf{Machine Learning Approach:} Models are implemented using the PyCaret framework, which provides automated preprocessing, model comparison, and evaluation. The evaluated models include Naive Bayes, Logistic Regression, and Random Forest.

\textbf{Deep Learning Approach:} The LSTM model is trained using the Adam optimizer and cross-entropy loss function. The training process is performed over multiple epochs with a fixed batch size to ensure stable learning.

\textbf{Imbalance Dataset:} Since negative reviews dominate, the model performance is evaluated carefully to ensure that minority classes are not ignored. This setup allows a fair comparison between Machine Learning and Deep Learning models in handling sentiment classification tasks.

\textbf{Experimental Consistency:} The training process is conducted using the same configuration across all runs. This ensures that the comparison between Machine Learning and Deep Learning approaches remains fair and reliable.

\setcounter{section}{8} 

\section{Results and Discussion}

\subsection{Machine Learning Model Results}
Machine Learning models were evaluated using PyCaret with TF-IDF feature representation. Among the three models evaluated, Logistic Regression demonstrated the most balanced performance across all metrics, achieving the highest F1-score and maintaining strong accuracy and recall values. Random Forest achieved competitive results but with slightly lower precision compared to Logistic Regression. Naive Bayes showed significantly lower performance, particularly in accuracy and recall, indicating its limitations in handling the lexical complexity and class imbalance of the dataset.

\subsection{LSTM Model Results}
The LSTM model was trained using preprocessed text data with a vocabulary derived from the training set. Training was conducted for 20 epochs using the Adam optimizer with a learning rate of 0.001 and cross-entropy loss. The model achieved a training accuracy of approximately 95\% and a test accuracy of 92\%, with a weighted F1-score of 91\%. The loss values decreased steadily across epochs, indicating effective convergence and learning of data patterns.

\subsection{Comparative Evaluation}
The evaluation results on the test set are presented in Table~\ref{tab:eval_metrics}. The LSTM model outperformed all Machine Learning baselines across all metrics. This improvement is attributed to LSTM's ability to capture sequential dependencies and contextual relationships within text, which classical models based on TF-IDF cannot achieve. However, the imbalanced nature of the dataset --- where negative reviews account for 67.47\% of samples --- may still influence classification performance for minority classes such as Neutral.

\begin{table}[h!]
\centering
\caption{Evaluation Metrics for Sentiment Classification}
\begin{tabular}{|l|c|c|c|c|}
\hline
\textbf{Model} & \textbf{Accuracy} & \textbf{Precision} & \textbf{Recall} & \textbf{F1-Score} \\ \hline
Naive Bayes & 0.4302 & 0.5660 & 0.4302 & 0.4767 \\ \hline
Logistic Regression & 0.7740 & 0.7242 & 0.7740 & 0.7311 \\ \hline
Random Forest & 0.7539 & 0.6972 & 0.7539 & 0.6963 \\ \hline
LSTM & 0.92 & 0.94 & 0.96 & 0.91 \\ \hline
\end{tabular}
\label{tab:eval_metrics}
\end{table}

As shown in Table~\ref{tab:eval_metrics}, the LSTM model achieves the highest performance across all four evaluation metrics. Logistic Regression is the best-performing classical ML model, which suggests that linear decision boundaries can still capture a reasonable degree of sentiment information when combined with TF-IDF. The gap between ML and DL approaches highlights the advantage of sequential modelling for informal, user-generated text data.
\subsection{Visualization of Model Performance}

\begin{figure}[t]
\centering
\begin{subfigure}[t]{0.49\linewidth}
    \centering
    \includegraphics[width=\linewidth]{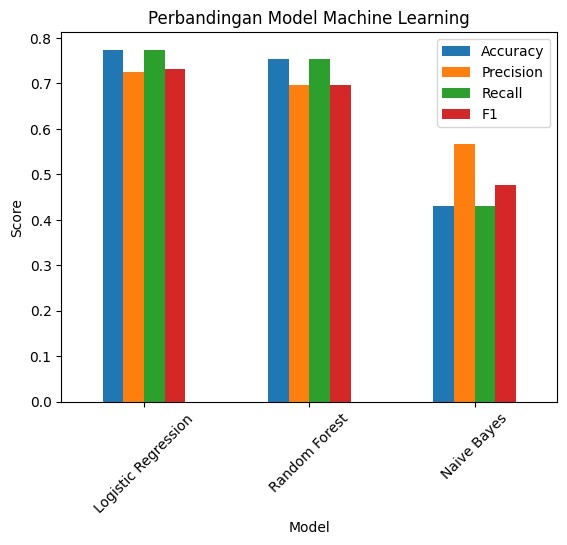}
    \caption{Machine Learning model comparison.}
    \label{fig:ml_comparison}
\end{subfigure}\hfill
\begin{subfigure}[t]{0.49\linewidth}
    \centering
    \includegraphics[width=\linewidth]{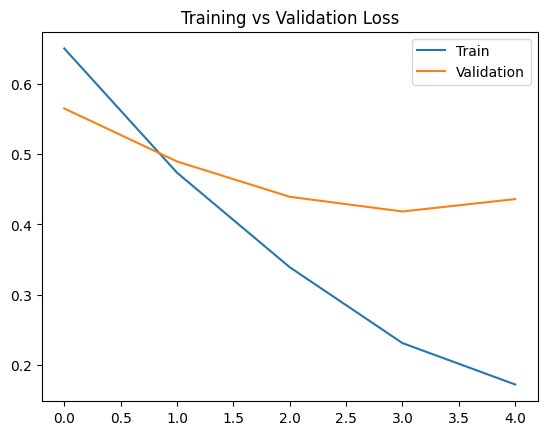}
    \caption{LSTM training and validation loss.}
    \label{fig:lstm_loss}
\end{subfigure}

\vspace{4pt}
\begin{subfigure}[t]{0.49\linewidth}
    \centering
    \includegraphics[width=0.9\linewidth]{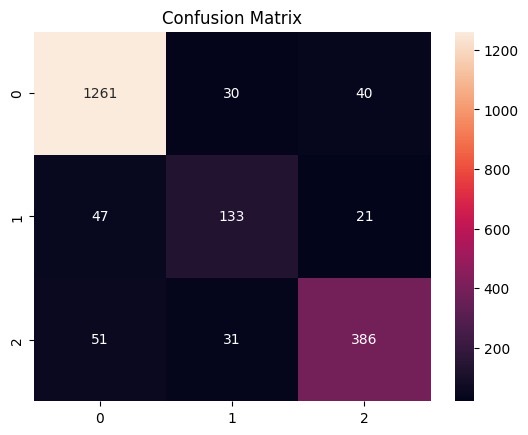}
    \caption{LSTM confusion matrix.}
    \label{fig:conf_matrix}
\end{subfigure}
\caption{Compact visualization of the evaluated models and LSTM behaviour.}
\label{fig:compact_results}
\end{figure}

To provide a clearer comparison of model performance across different evaluation metrics, a visual representation is presented in Figure~\ref{fig:ml_comparison}. This figure illustrates the performance of the Machine Learning models Logistic Regression, Random Forest, and Naive Bayes across four key metrics: Accuracy, Precision, Recall, and F1-score.

As shown in Figure~\ref{fig:ml_comparison}, Logistic Regression demonstrates the best overall performance among the Machine Learning models, achieving the highest accuracy and F1-score. Random Forest follows closely with competitive results, while Naive Bayes shows significantly lower performance across all evaluation metrics. These results indicate that Logistic Regression is more effective in capturing relevant patterns in the dataset compared to other traditional models.

To further analyze the training behavior of the Deep Learning model, the training and validation loss curves of the LSTM model are presented in Figure~\ref{fig:lstm_loss}. These curves provide insight into how the model learns from the training data and generalizes to unseen validation data over multiple epochs.

As illustrated in Figure~\ref{fig:lstm_loss}, the training loss decreases consistently across epochs, indicating that the model effectively learns patterns from the training data. The validation loss also shows an overall decreasing trend, although a slight increase is observed in the final epoch. This behavior suggests the presence of mild overfitting; however, the gap between training and validation loss remains relatively small. Therefore, the model can still be considered to generalize well, and the training process remains stable overall.

To further evaluate the classification performance of the LSTM model, a confusion matrix is presented in Figure~\ref{fig:conf_matrix}. This visualization provides a detailed breakdown of prediction results across all sentiment classes.

As shown in Figure~\ref{fig:conf_matrix}, the model demonstrates strong performance in correctly classifying the negative sentiment class, with a high number of true negative predictions. The positive class is also classified effectively, although a small number of instances are misclassified as negative. In contrast, the neutral class shows relatively lower classification performance, with several instances being incorrectly predicted as either negative or positive. This pattern indicates that the model has difficulty distinguishing neutral sentiment, which is likely influenced by the imbalanced distribution of the dataset and the inherent ambiguity of neutral expressions. Overall, the confusion matrix confirms that the model performs well on dominant classes while highlighting challenges in minority class classification.

\FloatBarrier
\subsection{Discussion}
The results confirm that LSTM-based deep learning is more effective than traditional machine learning for sentiment classification of Indonesian mobile game reviews. The informal linguistic characteristics of user reviews --- including slang, abbreviations, and mixed-language expressions --- benefit from LSTM's contextual understanding. However, the data imbalance problem remains a challenge. Future work should explore data balancing techniques such as Synthetic Minority Over-sampling Technique (SMOTE) or class weighting, as well as more advanced architectures such as Bidirectional LSTM (Bi-LSTM) or Transformer-based models like IndoBERT, to further improve performance on minority classes.

%% file: know.tex
\vspace{-10pt}
\section*{Acknowledgments}
\footnotesize
This project was completed as a final assignment for the Natural Language Processing course (Pengolahan Bahasa Alami, PBA 2026) at Institut Teknologi Sumatera (ITERA), Indonesia, by Group 16. The authors thank the dataset provider for making the Mobile Legends app review data publicly available on Kaggle, which served as the foundation for the sentiment analysis experiments in this study. The authors also acknowledge that parts of the project implementation and code preparation were assisted by an AI assistant named ChatGPT, PrismAI.

%% file: references.tex
\vspace{-12pt}
\section*{References}
\scriptsize

Setiawan, B. (2025).
A review of sentiment analysis applications in Indonesia between 2023--2024.
\textit{Journal of Information Engineering and Educational Technology}, 8(2), 71--83.
doi:10.26740/jieet.v8n2.p71-83.

Hamid, R. B., Andriansyah, C., Sensuse, D. I., Lusa, S., Elisabeth, D., \& Safitri, N. (2025).
Sentiment analysis and topic modeling for discovering knowledge in Indonesian mobile government applications.
\textit{Jurnal Teknik Informatika (JUTIF)}, 6(5), 3188--3203.
doi:10.52436/1.jutif.2025.6.6.4991.

Setyani, T., Sari, K., Heidy, H. N., \& Suryono, R. R. (2026).
Analisis sentimen pengguna aplikasi Jamsostek Mobile berdasarkan ulasan Google Play Store menggunakan algoritma support vector machine dan Naive Bayes.
\textit{MALCOM: Indonesian Journal of Machine Learning and Computer Science}, 6(1), 373--384.
doi:10.57152/malcom.v6i1.2526.

Ningsih, T. S., Hermanto, T. I., \& Nugroho, I. M. (2024).
Sentiment analysis of mobile provider application reviews using Naïve Bayes and support vector machine.
\textit{Sinkron: Jurnal dan Penelitian Teknik Informatika}, 8(2).
doi:10.33395/sinkron.v8i2.13469.

Tarwoto, T., Nugroho, R., Azka, N., \& Graha, W. S. R. (2025).
Analisis sentimen ulasan aplikasi Mobile JKN di Google Play Store menggunakan IndoBERT.
\textit{Jurnal Teknologi Informasi dan Komunikasi}, 9(2), 495--505.
doi:10.35870/jtik.v9i2.3340.

Pangestu, A. S., Christian, E., \& Lestari, A. (2025).
Aspect-based sentiment analysis pada ulasan pengguna aplikasi Mobile JKN menggunakan model berbasis transformer.
\textit{Journal of Information Technology and Computer Science}, 5(4).
doi:10.47111/jointecoms.v5i4.25348.

Putra, D. N. A. (2024).
Sentiment analysis of national health security mobile application review using machine learning.
\textit{Jurnal Jaminan Kesehatan Nasional}, 4(2).
doi:10.53756/jjkn.v4i2.269.

Fikri, A. A. Z., \& Ridho, H. (2025).
Identification of inconsistent reviews and ratings on apps using sentiment analysis: Case study on Indonesian digital media platform.
\textit{Metris: Jurnal Sains dan Teknologi}, 26(1).
doi:10.25170/metris.v26i01.6779.

Ayomi, J. M., Vitianingsih, A. V., Kristyawan, Y., Maukar, A. L., \& Widiartin, T. (2026).
Sentiment analysis of user reviews for the PLN Mobile application using Naïve Bayes and long short-term memory.
\textit{Journal of Information Systems and Informatics}, 7(4).
doi:10.63158/journalisi.v7i4.1342.

Nugroho, K. S., Sukmadewa, A. Y., Wuswilahaken, H. D. W., Bachtiar, F. A., \& Yudistira, N. (2021).
BERT fine-tuning for sentiment analysis on Indonesian mobile apps reviews.
\textit{arXiv preprint}.
\url{https://arxiv.org/abs/2107.06802}.